\documentclass{article}
\usepackage{amsmath,amsfonts}
\usepackage{algorithmic}
\usepackage{algorithm}
\usepackage{array}
\usepackage[caption=false,font=normalsize,labelfont=sf,textfont=sf]{subfig}
\usepackage{textcomp}
\usepackage{stfloats}
\usepackage{url}
\usepackage{verbatim}
\usepackage{graphicx}
\usepackage{cite}
\usepackage{xcolor}         % colors
\usepackage{multirow}
\usepackage{multicol}
\usepackage{graphics}
\usepackage{colortbl}
\usepackage{amsmath}
\usepackage{amssymb}
\usepackage{booktabs}
\usepackage{bbding}

\usepackage[preprint]{corl_2025} % Uncomment for pre-prints (e.g., arxiv); This is like ``final'', but will remove the CORL footnote.

\title{RGB-D Video Generation for Improving Human-to-Robot Object Handover Prediction}

\author{
  Tianyu Sun$^1$, Zhoujie Fu$^1$, Zihui Gao$^2$, Bang Zhang$^3$, Guosheng Lin$^1$\\
  $^1$College of Computing and Data Science, Nanyang Technological University\\
  $^2$College of Computer Science and Technology, Zhejiang University \\
  $^3$Alibaba Group\\
}

\begin{document}
\maketitle

%===============================================================================

\begin{abstract}
   Human-to-robot (H2R) object handover is a fundamental capability for human-robot collaboration, yet progress is hindered by the scarcity of large-scale, human-centric datasets and the significant sim-to-real gap. To address these challenges, we introduce Hand2Bot, an RGB-D video dataset that provides rich contextual information such as body posture and facial expressions, specifically collected for handover scenarios with real-world noise patterns. We further propose PassGen, a generative pipeline that leverages stable video diffusion and an Intention-Aware Temporal Face Encoder to synthesize realistic handover sequences while ensuring hand-object consistency. To bridge the sim-to-real gap, we implement a morphology-based depth editing strategy that replicates realistic sensor noise found in physical depth maps.  Experimental evaluations demonstrate that our framework achieves high intention identification accuracy and low false trigger rates in both ablation studies and real-world deployment on a physical robot platform. Our results confirm that training on PassGen allows for robust zero-shot transfer and earlier intention anticipation compared to traditional hand-centric baselines, effectively enabling socially aware robotic behavior in shared workspaces.
\end{abstract}


%===============================================================================

\section{Introduction}
Human-to-robot (H2R) object handover is one of the most fundamental tasks in robotics, serving as a cornerstone for enabling effective human–robot collaboration. Applications span assistive robots for the elderly, service robots in household environments, and collaborative systems on industrial assembly lines~\cite{yang2021reactive,ortenzi2021object}. The ability to safely and intuitively transfer objects between humans and robots not only enhances task efficiency but also serves as a critical precursor for complex downstream manipulation tasks. Consequently, proactively anticipating human communicative intent has attracted substantial attention in recent years~\cite{meng2022fast,christen2024synh2r,huang2024human}.

\begin{figure}[t]
    \centering
    \includegraphics[width=0.6\textwidth]{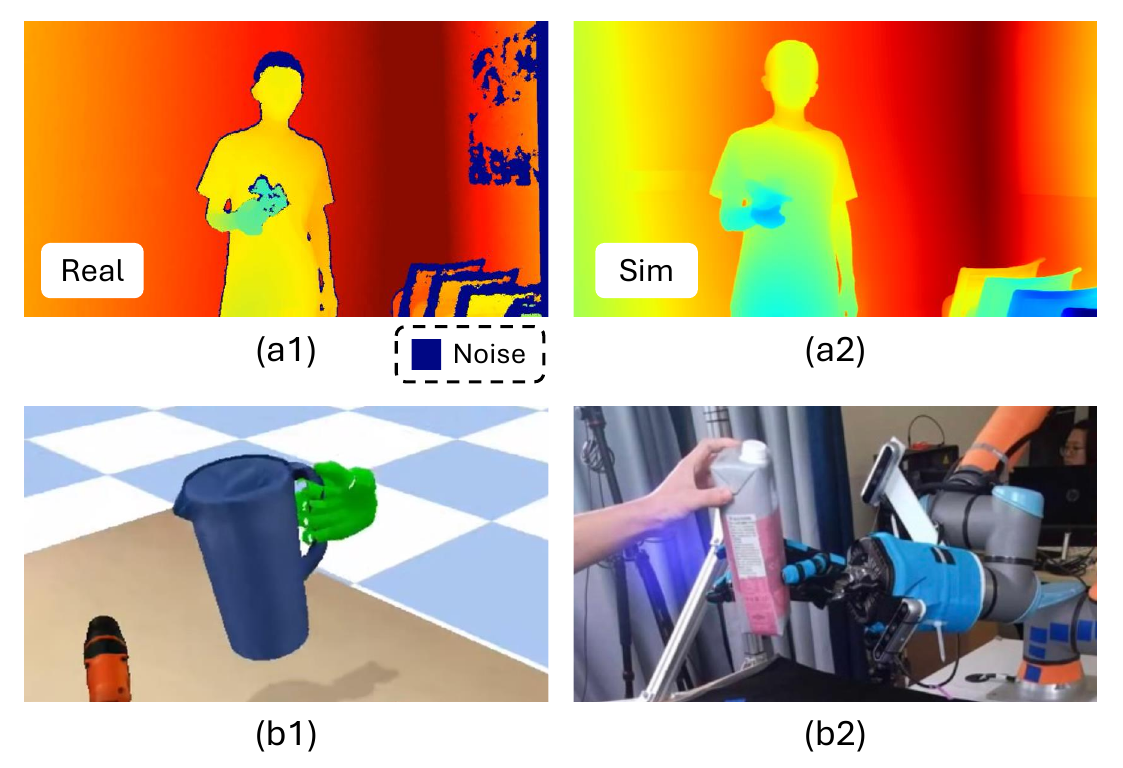}
    \caption{\textbf{Existing bottlenecks in current H2R handover datasets.} (a1) and (a2) compare a noisy real-world depth map against a clean synthetic depth map. (b1) and (b2) demonstrate the restricted, hand-centric field of view in mainstream benchmarks.}
    \label{fig:intro}
\end{figure}

Despite recent progress, existing datasets for H2R handover remain limited in their ability to support robust and generalizable learning due to two critical missing elements. First, the majority of available data is synthetic~\cite{chao2022handoversim,wang2024genh2r}, which fails to capture the complex noise characteristics of real-world sensors, including edge scattering, lighting variations, and motion blur~\cite{li2025segment}. As illustrated in Fig.~\ref{fig:intro} (a1) and (a2), raw physical depth observations retain severe stochastically distributed noise patterns; since real-time depth inpainting is computationally prohibitive for reactive control, models trained on clean synthetic geometry suffer from a pronounced simulation-to-reality (sim-to-real) gap. Second, mainstream benchmarks~\cite{wang2024genh2r,wang2025dexh2r} predominantly focus on egocentric or hand-centric views (Fig.~\ref{fig:intro}(b1), (b2)), completely omitting the broader body context. For natural and timely intention prediction, however, global human features—such as facial expressions, gaze direction, head orientation, and upper-body posture—are essential precursors to anticipate \textit{when} and \textit{how} a handover will be initiated~\cite{huang2025intention}. 

To bridge these gaps, our primary motivation is to construct a comprehensive dataset that captures the full-body context alongside physical sensor artifacts. However, scaling such real-world multi-modal data through manual capture incurs prohibitive annotation and recording overhead. This necessitates an algorithmic solution capable of synthesizing high-fidelity, interactive RGB-D sequences. Unfortunately, state-of-the-art generative models fall short in this specialized domain. General human video generation tasks rarely encounter human-object interactions from robot-perspective viewpoints~\cite{qiu2025lhm,mascaro2023hoi4abot}. More importantly, existing depth synthesis pipelines focus strictly on visual reconstruction fidelity, producing overly smoothed depth maps that lack the characteristic void patterns of standard sensors~\cite{yang2024depth,Hu2025depthcrafter}.

Driven by these algorithmic limitations, our core modification strategy introduces a unified generative and control framework. For the RGB modality, we propose \textbf{PassGen}, which builds upon video diffusion by embedding an Intention-Aware Temporal Face Encoder (TFE) to capture high-frequency facial micro-expressions and gaze focus. For the depth modality, we implement a morphology-based edge erosion strategy that directly models the realistic void distributions of an Intel RealSense L515 sensor. These generated streams are integrated with our real-world captures to construct the \textbf{Hand2Bot} dataset. Furthermore, to explicitly isolate perception enhancements from downstream manipulation constraints, we deploy an independent Intention Gating (IG) mechanism that leverages these global social cues to proactively trigger grasp trajectories.

Through extensive physical robot trials on a UR5e platform, we demonstrate that training on PassGen-augmented data yields exceptional zero-shot transfer capabilities. By shifting the tracking focus from localized hand tracking to multi-modal body-gaze context, the robot successfully eliminates false triggers during ambient human motion while achieving significantly earlier intent anticipation. 

In summary, our contributions are threefold:
\begin{itemize}
    \item We propose \textbf{PassGen}, a generative RGB pipeline implementing Stable Video Diffusion with a dedicated Temporal Face Encoder to maintain hand-object consistency and explicit social gaze cues.
    \item We introduce a morphology-based depth noise simulation that replicates realistic sensor-specific artifacts, effectively mitigating the sim-to-real gap.
    \item We construct the \textbf{Hand2Bot} dataset, providing 5,000 multimodal video pairs with rich full-body annotations, and validate its downstream robotic utility through extensive control and ablation experiments.
\end{itemize}

\section{RELATED WORKS}

\subsection{Human Video Generation for Interactive Scenarios}

Diffusion models have greatly pushed the boundary of generative research topics~\cite{yi2024mvgamba,sun2024diffusion}. Recent advances in human video generation~\cite{hu2024animate,sun2026mvanimateenhancingcharacteranimation} have begun to explore interactive scenarios where humans engage with objects or environments. Hu et al.~\cite{hu2025animate} propose Animate Anyone 2, which extends character animation pipelines to generate highly realistic human videos from reference images, with improvements in pose fidelity and motion consistency. Similarly, Xu et al.~\cite{xu2024anchorcrafter} introduce AnchorCrafter, a framework that allows controllable video synthesis conditioned on human-object interactions (HOI), enabling plausible motion dynamics when anchors such as objects are present. While these methods achieve impressive realism, they often focus primarily on pose or appearance consistency and lack explicit modeling of fine-grained human-object contact, which can lead to artifacts or implausible interactions in manipulation-heavy scenarios. Moreover, most existing approaches generate only RGB videos, without addressing the depth modality that is critical for robotic applications such as object handover. These limitations highlight the need for approaches that can robustly handle interactive human-object dynamics while producing RGB-D representations that bridge the gap between visual realism and robotic utility.

\subsection{Object Handover Datasets}

Several datasets have been proposed to study object handover, but each has notable limitations. HandoverSim~\cite{chao2022handoversim} and GenH2R~\cite{wang2024genh2r} are synthetic datasets designed to model H2R handover scenarios, which provide large-scale, controllable data but inevitably suffer from a sim-to-real gap due to the lack of realistic noise and variability found in real-world captures. On the other hand, DexH2R~\cite{wang2025dexh2r} includes real-world handover data, but it is primarily restricted to the hand regions, offering limited context about the human body or the surrounding environment that is crucial for modeling naturalistic interactions. Beyond explicit handover datasets, research works such as the subset of VidHOI implemented in HOI4ABOT~\cite{mascaro2023hoi4abot}, and the dataset constructed by Pang et al.~\cite{pang2025manivideo} include more parts of human bodies in the field of view for HOI scenarios. They provide diverse object manipulation scenes, but they do not specifically target handover tasks, which limits their applicability to robotic perception and control. These gaps highlight the need for a dataset that is both real-world and human-centric, as well as multimodal (RGB-D), specifically tailored for handover scenarios to support downstream robotic applications.

\section{METHODOLOGY}
\label{sec:method}

\begin{figure*}[t]
    \centering
    \includegraphics[width=\textwidth]{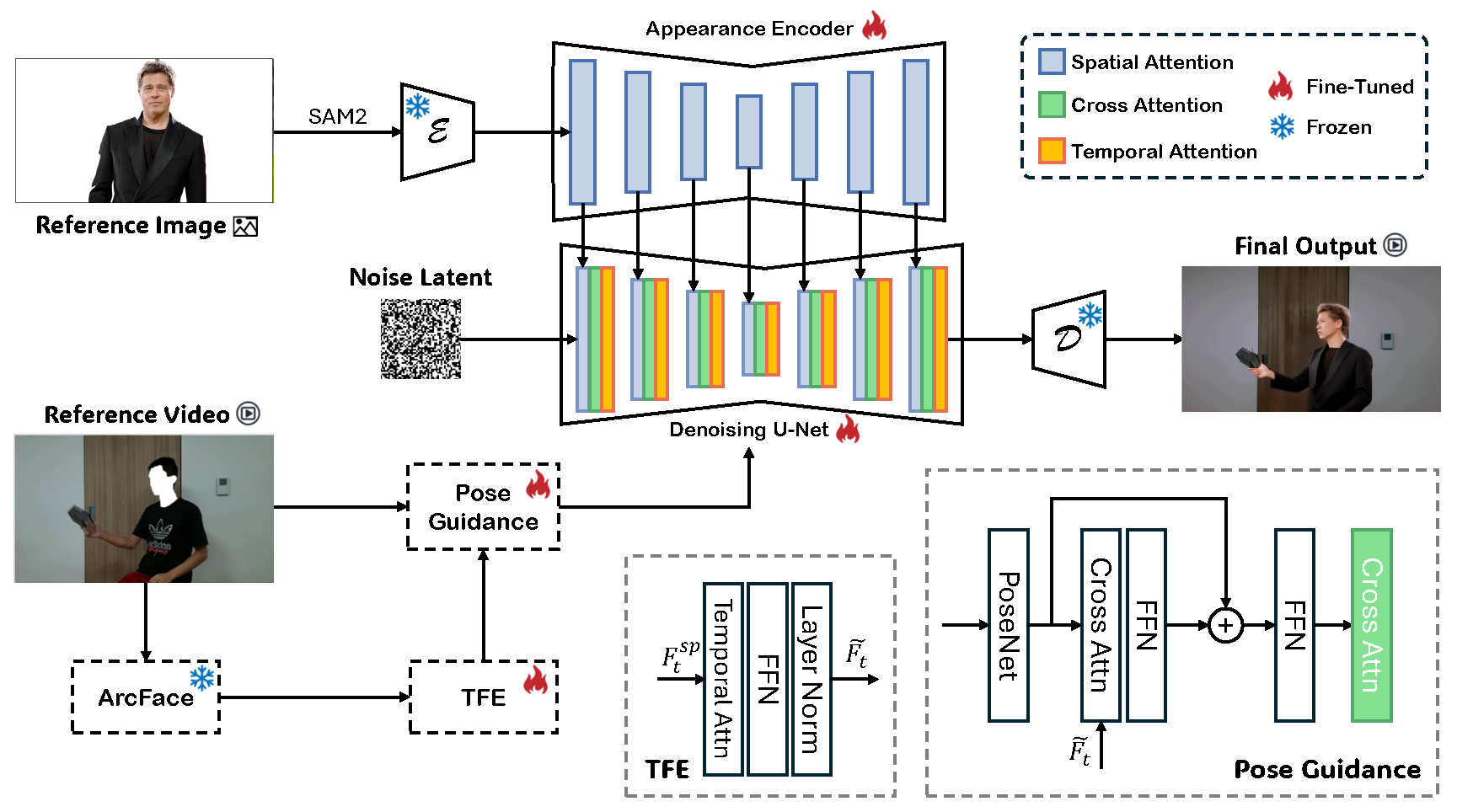}
    \caption{\textbf{Pipeline of our PassGen.} Our method is based on an SVD backbone, with two branches of networks. Given a pair of reference images $I_{ref}$ and video $V_{ref}$, the appearance feature from $I_{ref}$ and the temporal feature from $V_{ref}$ are operated by corresponding modules separately. We also elaborate on the structures of TFE and Pose Guidance (PG) blocks in the bottom right corner. }
    \label{fig:passgen}
\end{figure*}

Our goal is to generate realistic human-to-robot handover RGB-D videos that augment training data for downstream robotic tasks. To this end, we propose \textbf{PassGen}, a two-stage generation pipeline (Fig.~\ref{fig:passgen}) that first synthesizes human-object interaction RGB videos and then reconstructs depth maps with realistic noise patterns.

\subsection{Stage I: Pose-Guided RGB Video Generation}
We adopt a diffusion-based backbone based on Stable Video Diffusion (SVD)~\cite{blattmann2023stable}. To ensure the generated content is physically grounded for robotics, we condition the denoising U-Net on reference identity and precise motion sequences.

\textbf{1) Control Signal Integration:} Unlike text-driven models that may introduce semantic ambiguity, PassGen relies on deterministic physical signals. We utilize an Appearance Encoder (ReferenceNet structure~\cite{hu2024animate}) to extract identity features from a reference image $I_{ref}$, and a lightweight PoseNet~\cite{zhangmimicmotion} to process per-frame skeletons $V_{ref}$ extracted via DWPose~\cite{dwpose}. This multi-source conditioning constrains the latent space to adhere to the physical laws of human-object interaction.

\textbf{2) Intention-Aware Temporal Face Encoder (TFE):} To capture subtle handover intentions, we augment the pose guidance with a dedicated TFE branch. The design of TFE is theoretically motivated by the fact that in H2R scenarios, facial micro-expressions and gaze direction often precede physical reaching motions as precursor social signals. 

We extract facial embeddings using ArcFace~\cite{deng2019arcface} and process them through cascaded Temporal Attention (TA) and Feed Forward (FFN) blocks. Unlike standard animation models that treat the head as a rigid component, TFE models the facial region as a high-frequency semantic stream. This produces temporally-consistent face tokens $\tilde{F}_t$ that preserve the giver's visual focus, which is essential for the downstream robot to differentiate between an intentional handover and an ambient motion~\cite{mascaro2023hoi4abot}.

\textbf{3) Denoising and Training:} We fine-tune the SVD backbone using the LoRA scheme on our \textit{Hand2Bot-Real} dataset~\cite{hu2022lora}. The final pose-facial guidance $\tilde{P}_t$, which integrates both structural skeletons and social face tokens, is injected into the U-Net via cross-attention layers~\cite{animate-x,hu2025animate}:
\begin{equation}
\label{eq:ca}
U_t^l \leftarrow U_t^l + CrossAttention (Q=U_t^l, K=\tilde P_t, V=\tilde P_t)
\end{equation}
This ensures that every generated frame is not only spatially aligned with the handover trajectory but also contextually validated by explicit human social intention.

\subsection{Stage II: Morphological Depth Video Generation}
To bridge the sim-to-real gap, we simulate the specific noise characteristics of the \textbf{Intel RealSense L515} sensor, which often suffers from edge scattering and surface absorption.

\textbf{1) Depth Initialization:} We generate an initial affine-invariant depth map $D_{init}$ using DepthCrafter~\cite{Hu2025depthcrafter}. This ensures temporal smoothness across the handover sequence while maintaining a unified scale. We then perform a height-based linear alignment to substitute absolute depth values into the original human mask.

% \textbf{2) Realistic Noise Simulation:} Real-world depth data often contains ``void'' pixels (zero-depth). We record the noise distribution $N_0$ from our real-world collection and implement a \textbf{morphological erosion} strategy. We map boundary width vectors $w(h)$ relative to the pixel height $h$ to erode the generated depth $D_{gen}$. This process replicates the non-linearly distributed edge noise found in real L515 captures, producing the final realistic depth output $D_{final}$ with inter-frame alignment.

\textbf{2) Realistic Noise Simulation:} Real-world depth data often contains ``void'' pixels (zero-depth). We record the noise distribution $N_0$ from our real-world collection and implement a \textbf{morphological erosion} strategy. We map boundary width vectors $w(h)$ relative to the pixel height $h$ to erode the generated depth $D_{gen}$. This process serves as an empirical, heuristic morphology-based approximation of boundary-level depth artifacts, rather than a mathematically validated model of physical L515 sensor-noise distributions. This produces the final realistic depth output $D_{final}$ with inter-frame alignment.

By establishing physical consistency across the synthesized RGB-D streaming sequences, the \textit{PassGen} framework eliminates the ideal-geometry bias typical of synthetic engines, providing structurally grounded and sensor-realistic augmented data to train downstream workflows.

%%%%%%%%%%%%%%%%%%%%%%%%%%%%%%%%%%%%%%%%%%%%%%%%%%%%%%%%%%%%%%%%%%%%%%%%%%%%%%%%
\section{INTENTION-GATED PERCEPTION AND CONTROL SYSTEM}
\label{sec:control_system}

% To transition from passive, reactive grasping to socially cooperative interaction, we introduce an independent Intention Gating (IG) mechanism. This module operates as a high-level semantic filter that bridges multimodal human context perception with downstream robotic grasp execution, isolating intent detection from physical manipulation constraints.

To transition from passive, reactive grasping to socially cooperative interaction, we introduce an independent Intention Gating (IG) mechanism. This module operates as a heuristic multimodal gating mechanism that bridges multimodal human context perception with downstream robotic grasp execution, isolating intent detection from physical manipulation constraints.

\subsection{Physical Camera Configuration and Application Scenario}
As illustrated in our scenario definition, capturing full-body cues while maintaining reliable facial gaze tracking poses a coupled perception challenge. To resolve this, our application environment deploys the Intel RealSense L515 sensor at a calibrated robot-centric perspective: positioned on the UR5e robot base, tilted upwards facing the interaction zone. This geometric configuration ensures that the camera’s wide field-of-view encompasses the participant's upper body and arm kinematics (from a distance of $1.5\,\text{m}$ to $2.2\,\text{m}$), while providing sufficient facial spatial resolution to resolve high-frequency micro-expressions and gaze direction stream inputs.

\subsection{Multimodal Fusion Formulation}
The gating logic continuously ingests synchronized multi-modal features to compute a unified scalar intention score $S_{intent}(t) \in [0, 1]$:
\begin{equation}
\label{eq:gating_score}
S_{intent}(t) = \sigma \left( w_g \cdot f_{gaze}(\tilde{F}_{t}) + w_v \cdot v_{obj}(t) \right)
\end{equation}
where $f_{gaze}(\tilde{F}_{t})$ represents the temporal gaze focus confidence score extracted via the Temporal Face Encoder (TFE), and $v_{obj}(t) = -\frac{\partial D_{obj}}{\partial t}$ denotes the physical approaching velocity computed from the masked depth boundaries. The weights $w_g$ and $w_v$ serve as scaling coefficients to balance the dimensionless gaze prediction with the metric velocity scalar.

\subsection{Monotone Constraints and Sensitivity Analysis}
During dynamic interaction, temporary gaze drops or localized occlusions can induce stochastic oscillations in instantaneous tracking outputs. To circumvent unstable state switching in physical deployments, we enforce a monotone non-decreasing constraint on the final inference trigger boundary:
\begin{equation}
\hat{S}_{intent}(t) = \max \left( \hat{S}_{intent}(t-1), \, S_{intent}(t) \right)
\end{equation}
The parallel-jaw gripper's grasp trajectory is proactively initiated if and only if $\hat{S}_{intent}(t) \ge \tau$, where $\tau$ is a strict safety activation threshold. 

To justify our parameterization, a sensitivity analysis was performed on the threshold boundary. We set $\tau = 0.80$ optimizes system performance, suppressing accidental activations from ambient user motion while minimizing the tracking latency before the presentation period terminates.

%%%%%%%%%%%%%%%%%%%%%%%%%%%%%%%%%%%%%%%%%%%%%%%%%%%%%%%%%%%%%%%%%%%%%%%%%%%%%%%%

\begin{table*}[t]
    \centering
    \caption{\textbf{Animation Results on Hand2Bot-Real Dataset.} \textcolor{red}{Red} marks the best algorithm, and \textcolor{blue}{blue} marks the second-best. }
    \begin{tabular}{l|cccc|cc}
        \hline
        \multirow{2}{*}{Method} & \multicolumn{4}{c}{\cellcolor{lightgray!30}Image} & \multicolumn{2}{|c}{\cellcolor{lightgray!30}Video}\\
        \cline{2-7}
        ~ & PSNR $\uparrow$ & SSIM $\uparrow$ & LPIPS $\downarrow$ & FID $\downarrow$ & FID-VID $\downarrow$ & FVD $\downarrow$\\
        \hline
        MagicPose~\cite{chang2024magicpose} & 15.71 & 0.733 & 0.412 & 243.70 & 295.61 & 714.36\\
        MagicAnimate~\cite{xu2024magicanimate} & 16.21 & 0.740 & 0.400 & 212.48 & 284.08 & 947.65\\
        AnimateAnyone~\cite{hu2024animate} & 16.51 & 0.749 & 0.343 & 190.02 & 203.31 & 623.79\\
        DreamPose~\cite{karras2023dreampose} & 16.58 & 0.753 & 0.352 & 173.90 & 192.13 & 682.55\\
        MimicMotion~\cite{zhangmimicmotion} & 19.48 & 0.780 & 0.340 & 111.39 & \textcolor{red}{97.70} & 439.65\\
        Animate-x~\cite{animate-x} & 22.23 & 0.853 & 0.235 & 130.25 & 142.47 & 568.99\\
        StableAnimator~\cite{tu2025stableanimator} & \textcolor{blue}{23.17} & \textcolor{blue}{0.867} & \textcolor{blue}{0.218} & \textcolor{blue}{101.67} & 106.73 & \textcolor{blue}{372.81}\\
         \hline
        PassGen & \textcolor{red}{25.12} & \textcolor{red}{0.909} & \textcolor{red}{0.212} & \textcolor{red}{91.32} & \textcolor{blue}{99.77} & \textcolor{red}{337.59}\\
         \hline
    \end{tabular}
    \label{tab:h2b}
\end{table*}

\begin{figure*}[t]
    \centering
    \includegraphics[width=\textwidth]{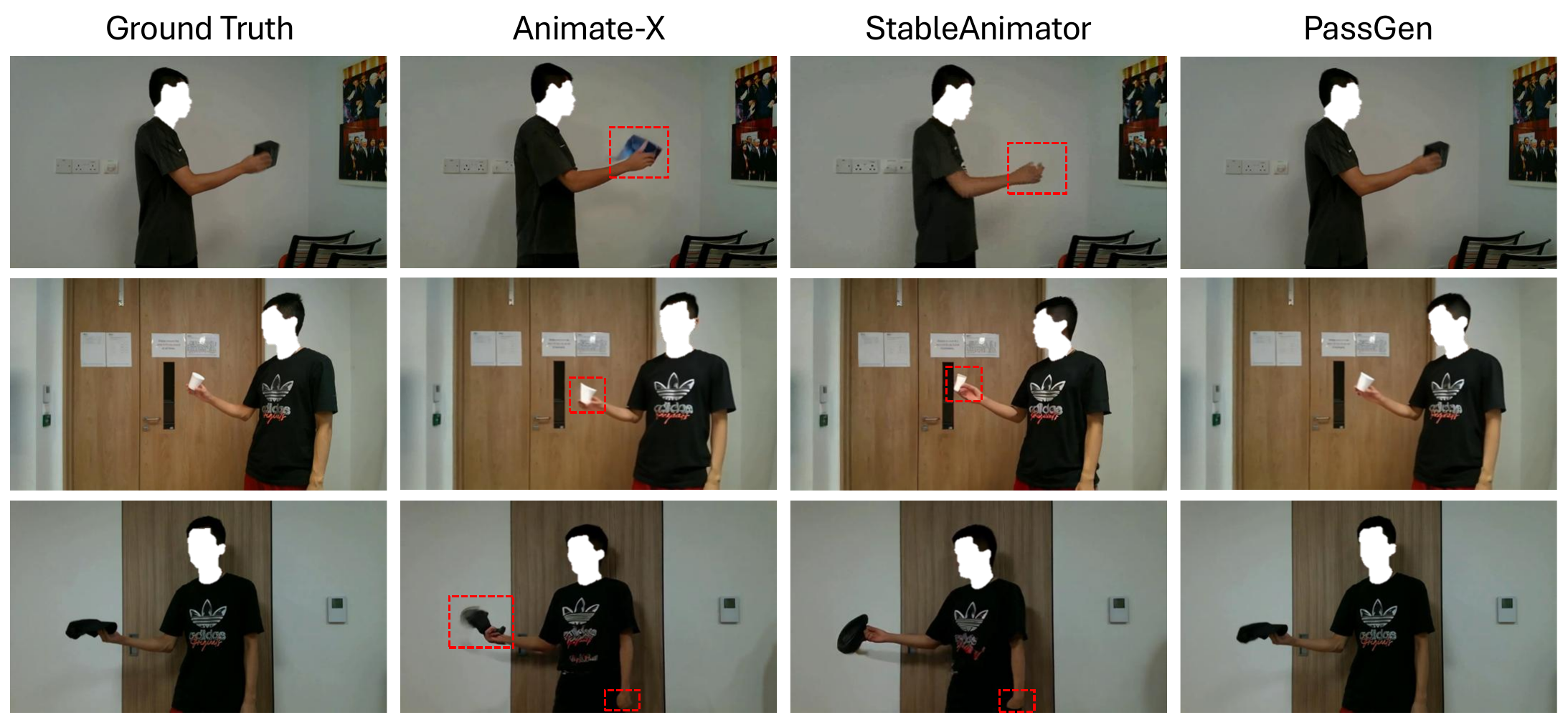}
    \caption{\textbf{Qualitative results on Hand2Bot-Real dataset.} We compare our method with other SOTA human video generation algorithms on different reference images and video frames from our Hand2Bot-Real dataset. Our PassGen shows a promising and stable performance compared with other SOTA methods. We mark the artifacts of the previous methods with \textcolor{red}{red} boxes. }
    \label{fig:h2b}
\end{figure*}

\section{EXPERIMENTS}
We evaluate PassGen on the Hand2Bot dataset across two dimensions: 1) the fidelity of the generated RGB-D sequences, and 2) its utility for downstream robotic intention sensing.

\subsection{Setup and Metrics}
The training and testing of PassGen is implemented on NVIDIA A6000. We assess visual quality using PSNR, SSIM, and LPIPS for frames, and FID-VID~\cite{balaji2019conditional} and FVD~\cite{unterthiner2018towards} for temporal coherence. 
Crucially, for robotics utility, we evaluate the Intention Identification Accuracy and False Positive Rate (FPR) of a handover policy trained on our data.

\subsection{Animation Results and Ablation}
We evaluate PassGen against state-of-the-art animation models on a test set of 250 clips from \textit{Hand2Bot-Real}. As shown in Tab.~\ref{tab:h2b}, PassGen consistently outperforms existing methods across both frame-level (PSNR, SSIM) and video-level (FVD, FID-VID) metrics, indicating superior spatial fidelity and temporal stability. 

Qualitative comparisons are presented in Fig.~\ref{fig:h2b}. While models such as StableAnimator~\cite{tu2025stableanimator} and MimicMotion~\cite{zhangmimicmotion} achieve high texture fidelity, they frequently struggle with hand-object consistency during rapid handover trajectories. Specifically, as highlighted in the zoomed-in regions of Fig.~\ref{fig:h2b}, prior methods often exhibit severe artifacts, such as object distortion or finger-mask flickering, when the human giver performs complex reaching motions. In contrast, PassGen maintains a stable interaction interface throughout the sequence. This is attributed to our Pose-Guidance and TFE modules, which effectively bridge the gap between global body dynamics and local manipulation cues, ensuring the generated RGB-D pairs plausible for downstream robotic training.

As for the ablation study, tab.~\ref{tab:abl} quantifies the contribution of our conditioning modules. Removing the Temporal Face Encoder (TFE) leads to a significant drop in FVD, confirming that gaze and facial consistency are vital for realistic human-to-robot interaction synthesis.

\begin{table*}[t]
    \centering
    \caption{\textbf{Ablation Study on PassGen Modules.} We compare our PassGen with the conditions without several crucial blocks, including ArcFace, Temporal Face Encoding (TFE), and Pose Guidance (PG), on the Hand2Bot-Real dataset test set. We take two image-based and two video-based metrics for comparison. We mark the best metrics in \textbf{bold}. }
    \begin{tabular}{cccc|cccc}
        \hline
        Backbone & Pose Guidance & TFE & ArcFace & FID-VID $\downarrow$ & FVD $\downarrow$ & PSNR $\uparrow$ & SSIM $\uparrow$\\
        \hline
        \checkmark & \checkmark & \checkmark & \checkmark & \textbf{99.77} & \textbf{337.59} & \textbf{25.12} & \textbf{0.909} \\
        \checkmark & \checkmark & \checkmark &  & 107.44 & 373.61 & 22.77 & 0.829\\
        \checkmark & \checkmark &  &  & 111.60 & 399.52 & 21.47 & 0.809\\
        \checkmark &  &  &  & 117.43 & 422.90 & 19.12 & 0.791\\
        \hline
    \end{tabular}
    \label{tab:abl}
\end{table*}

\subsection{Depth Fidelity Analysis}

\begin{figure*}[h]
    \centering
    \includegraphics[width=\textwidth]{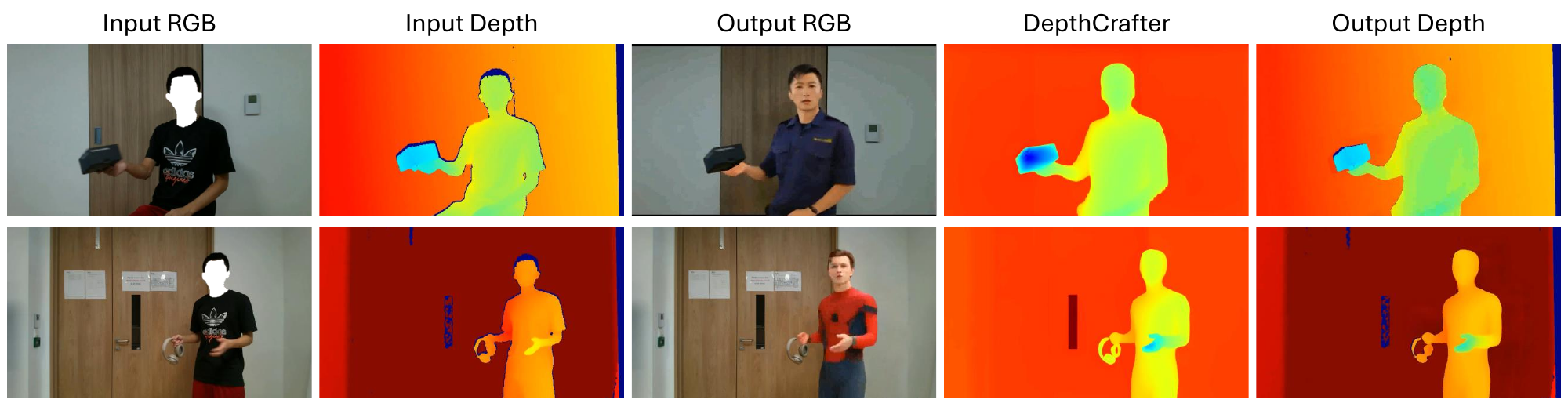}
    \caption{\textbf{Qualitative comparison of generative RGB-D sequences.} Our PassGen pipeline effectively reconstructs human-object interaction geometry from sparse L515 inputs while maintaining realistic sensor noise patterns compared to over-smoothed baselines like DepthCrafter~\cite{Hu2025depthcrafter}. }
    \label{fig:depth}
\end{figure*}

The primary objective of our depth generation is twofold: ensuring pixel-level alignment with the synthesized RGB video while preserving the inherent noise patterns of physical sensors. As shown in Fig.~\ref{fig:depth}, generic depth estimation (e.g., DepthCrafter~\cite{Hu2025depthcrafter}) produces edges that are far too ``clean'' for robotic perception training. Our approach intentionally reintroduces morphological artifacts at the human-object interface. This ensures that the generated depth sequence maintains the realistic format, which is critical for training robust policies that do not over-fit to perfect synthetic geometry.

\subsection{Real H2R Handover Evaluation}
To validate the practical utility of our proposed framework, we deployed the system on a physical UR5e platform. We conducted human-to-robot handover trials to assess the effectiveness of the \textit{PassGen}-augmented policy and the \textit{Intention Gating} (IG) scheme.

\subsubsection{Experimental Setup}
We selected 10 distinct objects, including 5 \textbf{Seen Objects} and 5 \textbf{General Unseen} objects. To evaluate the intention gating mechanism, we conducted:
% \begin{itemize}
%     \item \textbf{Positive Trials}: 10 intentional handover attempts per object type to evaluate the Intention Success Rate (ISR).
%     \item \textbf{Negative Trials}: unrelated movements to evaluate the False Trigger Rate (FTR).
% \end{itemize}

\begin{itemize}
    \item \textbf{Positive Trials}: 10 intentional handover attempts per object type to evaluate the Intention Success Rate (ISR). We explicitly define ISR as an intention-triggering metric that measures whether the robot correctly initiates a reaching action during the presentation period, rather than evaluating full mechanical grasp completion or physical manipulation generalization.
    \item \textbf{Negative Trials}: unrelated movements to evaluate the False Trigger Rate (FTR).
\end{itemize}

\subsubsection{Results Analysis}
The experimental results in Tab.~\ref{tab:real_world_results} demonstrate that the system maintains high reliability, particularly in identifying handover intent. Our full module achieves an overall Mean ISR of $54/60$, indicating robust performance across diverse object categories.

\begin{table}[h]
\centering
\caption{\textbf{Real-World Handover Performance}: intention reliability and safety evaluation on real-world scenarios.}
\label{tab:real_world_results}
\begin{tabular}{lcccc}
\toprule
\multirow{2}{*}{Object} & \multirow{2}{*}{Geometric Category} & \multicolumn{2}{c}{w IG} & w/o IG \\
\cmidrule(lr){3-4} \cmidrule(lr){5-5}
 & & \textbf{ISR} $\uparrow$ & \textbf{FTR} $\downarrow$ & \textbf{FTR} $\downarrow$ \\
\midrule
\multirow{5}{*}{Seen} 
 & Box        & 10 / 10 & 1 / 5  & 5 / 5 \\
 & Cylinder   & 10 / 10 & 0 / 5  & 5 / 5 \\
 & Cup        & 9 / 10 & 0 / 5  & 4 / 5 \\
 & Plate      & 10 / 10 & 0 / 5  & 5 / 5 \\
 & Irregular  & 8 / 10 & 1 / 5  & 3 / 5 \\
\cmidrule{1-5}
Unseen & \textit{-} & 7 / 10 & 0 / 5 & 3 / 5 \\
\midrule
\textbf{Overall} & & 54 / 60 & 2 / 30 & 25 / 30 \\
\bottomrule
\end{tabular}
\end{table}

\textbf{Analysis of Failures:} Performance degradation observed in the \textit{Irregular} and \textit{Unseen} categories (ISR of 80\% and 70\%, respectively) stems from two primary factors. First, the object term of the intention score occasionally fails with irregular geometries, as the system struggles to determine if the object has been moved forward into the interaction zone. Second, even when the intention is correctly identified, the complex surface of these objects makes them physically hard to grasp for the parallel-jaw gripper, leading to failed completions despite a correct trigger.

\textbf{Safety and Baseline Comparison:} The system achieved a low overall FTR of $2/30$ with the intention gating mechanism active, confirming that the gaze-aware module effectively prevents accidental triggers. In stark contrast, the \textit{Baseline} configuration (w/o IG) suffered from a severe false trigger rate of $25/30$, effectively failing to distinguish between intentional handovers and ambient movements. In negative trials, the robot remained stationary even when the hand was near the object, provided the human's visual focus was directed elsewhere. This significant reduction in FTR validates the safety and necessity of our approach in shared human-robot workspaces.

% We also provide representative cases for each object type and interaction scenario in the video supplementary material.

%%%%%%%%%%%%%%%%%%%%%%%%%%%%%%%%%%%%%%%%%%%%%%%%%%%%%%%%%%%%%%%%%%%%%%%%%%%%%%%%

\section{CONCLUSION}

In this work, we present \textbf{PassGen}, a novel algorithm for generating high-quality RGB-D human videos that can robustly handle H2R handover scenarios. Our method tackles key challenges, including intricate pose articulation for human video generation and data scarcity for H2R handover. We also construct the real-world Hand2Bot-Real dataset and augment it with PassGen to form the \textbf{Hand2Bot} dataset. Furthermore, the inclusion of realistic depth noise simulation proves essential for robust sim-to-real transfer, significantly improving performance on physical robot platforms. While the object term of the intention score occasionally fails with irregular geometries and certain objects remain physically hard to grasp, our gating mechanism effectively prevents accidental triggers. We anticipate that our approach and dataset will provide a valuable foundation for future studies in video generation, proactive H2R handover, and socially aware robotic perception.

% In this work, we present \textbf{PassGen}, a generative RGB-D human video pipeline that addresses pose articulation complexity and data scarcity in H2R handover scenarios. Combining real-world captures with PassGen synthesis, we construct the \textbf{Hand2Bot} dataset, incorporating morphological depth noise simulation to facilitate robust sim-to-real transfer on physical robot platforms. While irregular geometries and challenging grasp profiles can limit object-level completion, our safety-first gating mechanism effectively suppresses accidental triggers. We anticipate that our dataset and framework will serve as a practical foundation for future research in socially aware robotic perception and proactive handover control.

\newpage
\bibliography{example}  % .bib

\appendix
\section{HAND2BOT DATASET}
To address the scarcity of high-quality human-to-robot (H2R) handover data, we construct the Hand2Bot dataset. Unlike existing datasets, Hand2Bot captures the full human body field of view and incorporates realistic sensor noise, facilitating robust intention sensing and sim-to-real transfer.

\subsection{Construction and Annotation}
Hand2Bot contains 5,000 RGB-D video pairs (2,125 real-world; 2,875 generated via \textit{PassGen}). For the real subset, we utilized an Intel RealSense L515 camera across 7 diverse indoor scenes with 33 common household objects. 

% \textbf{Intention-Centric Labeling:} Critically, we include approximately 20\% of video pairs as negative cases where the human moves with the object but has no intention to hand it over. To support downstream timing, we provide timestamp-based labels for the \textit{presentation period}~\cite{ortenzi2021object}, marking the exact frames where handover intention is explicitly initiated.

\textbf{Intention-Centric Labeling:} Critically, we include approximately 20\% of video pairs as negative cases where the human moves with the object but has no intention to hand it over. In our dataset, positive handover cases comprise clear gaze-directed object approach motions. Negative cases extend beyond ambient background movements to include ambiguous gestures, weak-gaze reaching, and non-gaze-oriented tool passes, ensuring the evaluation is not biased toward scenarios artificially favorable to our gating cues. To support downstream timing, we provide timestamp-based labels for the \textit{presentation period}~\cite{ortenzi2021object}, marking the exact frames where handover intention is explicitly initiated.

% \textbf{Grasp Pose Labeling:} To support manipulation tasks, we integrate GraspNet~\cite{fang2020graspnet} to generate 6-DoF parallel-gripper grasp poses, with the workspace mask restrained to the object region. We implement a temporal consistency filter to select stable pose sequences during the presentation period, ensuring they are physically reachable for robotic executors. We keep the pose result that is not in contact with the human hand. The grasp pose we implement is the form of parallel grippers, as shown in the visualization results in Fig.~\ref{fig:pose}. 

\begin{figure}[h]
    \centering
    \includegraphics[width=0.6\textwidth]{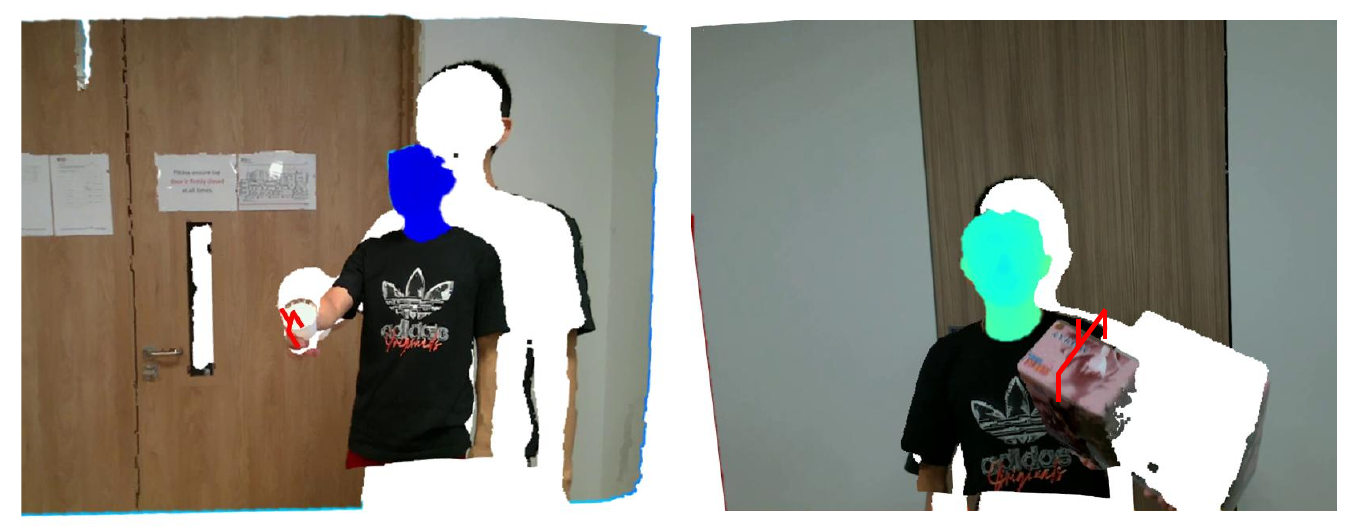}
    \caption{\textbf{Grasp Pose Annotation.} We give two examples of the GraspNet-based annotation on our Hand2Bot dataset. We mask all facial regions and replace them with the corresponding visualized depth maps. }
    \label{fig:pose}
\end{figure}

\textbf{Grasp Pose Labeling:} To support downstream manipulation tasks, we integrate GraspNet~\cite{fang2020graspnet} to generate a candidate pool of 6-DoF parallel-gripper grasp poses, with the workspace mask strictly constrained to the object region, as shown in the visualization results in Fig.~\ref{fig:pose}. To guarantee execution stability and physical reachability for the robotic executor, we implement a \textit{vertical-alignment filtering strategy} rather than relying on a singular fixed heuristic. 

Specifically, let $\mathbf{R} \in SO(3)$ represent the orientation matrix of a candidate grasp pose, and $\mathbf{v}_z = [0, 0, -1]^T$ denote the canonical downward vertical axis aligned with gravity. We evaluate the angular deviation $\theta = \arccos(\mathbf{g}_z \cdot \mathbf{v}_z)$ for all valid, collision-free candidate poses returned by GraspNet, where $\mathbf{g}_z$ is the approach vector of the gripper. The system dynamically selects the kinematically feasible pose that minimizes $\theta$, effectively prioritizing top-down vertical grasp trajectories while retaining the flexibility of full 6-DoF adaptation to arbitrary object geometries. We implement a temporal consistency filter to select stable pose sequences during the presentation period and exclude any pose profiles that establish physical contact with the human hand region.

\begin{table*}[h]
\centering
  \caption{\textbf{Statistical comparison.} \textit{Noise?} represents whether the depth information in the dataset considers real-world noise. \textit{Scope} stands for the field of view for the data. \textit{Paradigm} represents the task setting, where \textit{A} indicates that the giver/receiver can be anything. \textit{\#Realistic Video} is the number of realistic RGB(-D) videos. Note that the dataset name in \cite{laplaza2022context} is not specified. }
  \label{tab:stat}
  \resizebox{\textwidth}{!}{%
    \begin{tabular}{@{}cccccccccc@{}}
    \toprule
    \multirow{2}{*}{Dataset} & \multirow{2}{*}{Source} &
      \multicolumn{4}{c}{Data Composition} &
      \multicolumn{3}{c}{Handover Detail} &
      \multirow{2}{*}{\#Realistic Video} \\ 
    \cmidrule(lr){3-6} \cmidrule(lr){7-9}
    &  & RGB & Depth & Noise? & Scope & Paradigm & Giver & Receiver &  \\ 
    \midrule
    HandoverSim~\cite{chao2022handoversim} & Synthetic & \checkmark & \checkmark & \scalebox{0.85}[1]{$\times$} & Hand & H2R & Hand & Parallel Gripper & - \\
    GenH2R~\cite{wang2024genh2r} & Synthetic & \checkmark & \checkmark & \scalebox{0.85}[1]{$\times$} & Hand  & H2R & Hand & Parallel Gripper & - \\
    \cite{laplaza2022context} & Real & \checkmark & \checkmark & \checkmark & Human & H2R & Hand & -  & 330 \\
    DexYCB~\cite{chao2021dexycb} & Real & \checkmark & \checkmark & \checkmark & Hand & H2A & Hand & -  & 1,000 \\
    H2O~\cite{ye2021h2o}  & Real & \checkmark & \scalebox{0.85}[1]{$\times$} & - & Hand & H2H  & Hand  & Hand & 1,200 \\
    HOH~\cite{wiederhold2023hoh} & Real & \checkmark & \checkmark & \checkmark & Hand & H2H  & Hand & Hand & 2,720 \\ 
    DexH2R~\cite{wang2025dexh2r} & Real & \checkmark & \checkmark & \checkmark & Hand & H2R & Hand & ShadowHand & 4,282 \\ 
    \midrule
    Hand2Bot & Both & \checkmark & \checkmark & \checkmark & Human & H2R & Hand & Parallel Gripper & 5,000 \\ 
    \bottomrule
    \end{tabular}%
  }
\end{table*}

\subsection{Benchmark Comparison and Advantages}
As illustrated in Tab.~\ref{tab:stat}, Hand2Bot provides several unique advantages over existing benchmarks that are critical for real-world deployment. While prior synthetic datasets such as HandoverSim~\cite{chao2022handoversim} and GenH2R~\cite{wang2024genh2r} are noise-free and thus suffer from a pronounced sim-to-real gap, our inclusion of realistic L515 sensor noise patterns facilitates more robust policy transfer to physical systems. Furthermore, most real-world datasets~\cite{chao2021dexycb, ye2021h2o, wang2025dexh2r} predominantly utilize hand-centric views, which limit the perception system to local cues. In contrast, Hand2Bot offers a full-body field of view, enabling the exploitation of global semantic features—such as gaze direction, head orientation, and upper-body posture—which are essential for accurate human intention anticipation. Finally, by specifically targeting the H2R paradigm with a parallel-gripper configuration and providing the largest collection of realistic RGB-D clips to date, Hand2Bot serves as a comprehensive bridge between high-level intention sensing and low-level robotic manipulation.

All real-world data collection and experiments were conducted with the informed consent of the participants. We only focus on non-invasive motion and intention sensing. 

%%%%%%%%%%%%%%%%%%%%%%%%%%%%%%%%%%%%%%%%%%%%%%%%%%%%%%%%%%%%%%%%%%%%%%%%%%%%%%%%

\section{Supplemental Results}

\subsection{Intention Gating Performance and Ablation Study}
To evaluate the efficacy of our intention gating mechanism, we conduct an ablation study on the \textit{Intention Score Module} using an evaluation subset of the Hand2Bot-Real dataset. 

As summarized in Tab.~\ref{tab:ablation}, the baseline lacking both terms fails to reject ambient movements (95.5\% FPR). While the object term alone marginally improves accuracy (78.2\%), adding the gaze term significantly drives system robustness, dropping the FPR to 31.8\%. Our Full Module achieves the best overall performance (90.0\% Mean Acc., 13.6\% FPR), confirming that integrating visual attention with spatial reaching cues is essential for safe operation. For evaluation, a trial is labeled as a successful grasp based on strict safety constraints rather than fixed spatio-temporal thresholds: the gripper must establish stable force closure on the target object without any physical collision or proximity interference with the human hand.

% --- Table Section ---
\begin{table}[h]
\centering
\caption{\textbf{Ablation Study of Intention Score Module.} Performance comparison on the Hand2Bot-Real evaluation subset.}
\label{tab:ablation}
\begin{tabular}{lcc|cc}
\toprule
Configuration & Gaze Term & Object Term & Mean Acc. $\uparrow$ & FPR $\downarrow$ \\
\midrule
Baseline      &           &             & 72.7\%          & 95.5\%          \\
Gaze-only     & \Checkmark &             & 86.4\%          & 31.8\%          \\
Object-only   &           & \Checkmark   & 78.2\%          & 77.3\%          \\
\textbf{Full Module} & \Checkmark & \Checkmark & \textbf{90.0\%} & \textbf{13.6\%} \\
\bottomrule
\end{tabular}
\end{table}

\subsection{Downstream Utility: Real-Only vs. Augmented Training}
To establish the explicit value of our synthesized data and address potential generative bias concerns, we conducted a direct downstream comparison using an identical prediction architecture and evaluation setting. We trained the intention prediction model under two data configurations: (i) \textit{Real}, where the network was optimized solely on the real-world training video pairs, and (ii) \textit{Real + Syn}, where the network was trained on all training video pairs augmented by our generative pipeline.

\begin{table}[h]
\centering
\caption{\textbf{Ablation Study}: we compare the results w/o Hand2Bot-Syn}
\label{tab:real_vs_gen_ablation}
\begin{tabular}{lccc}
\toprule
Training Data & Mean Acc. $\uparrow$ & Mean FPR $\downarrow$ & Unseen Obj. ISR $\uparrow$ \\
\midrule
Real & 87.5\% & 22.8\% & 6/10 \\
\textbf{Real + Syn} & \textbf{90.0\%} & \textbf{13.6\%} & \textbf{7/10} \\
\bottomrule
\end{tabular}
\end{table}

As detailed in Table~\ref{tab:real_vs_gen_ablation}, augmenting training with 2,875 synthesized sequences from \textit{PassGen-Syn} significantly improves downstream tracking accuracy, reduces FPR, and enhances robustness against overfitting on unseen objects. This validates that our structural pose constraints and facial signal modeling successfully increase physical variability. Given the limited margin on unseen objects ($N=10$ trials) without formal statistical significance across multiple random seeds, we frame the synthetic augmentation primarily as a variance-reduction and false-positive suppression mechanism rather than a claim of broad manipulation superiority.

\subsection{Gating Sensitivity Analysis and Parameter Configuration}
\label{subsec:gating_sensitivity}

To evaluate the robustness of our Intention Gating (IG) mechanism and justify our design choices, we conducted an offline parameter analysis. The scalar intent score combines a dimensionless facial gaze confidence $f_{gaze} \in [0,1]$ and a metric spatial approach velocity $v_{obj}$ ($\text{m/s}$). To resolve this scale discrepancy, the fusion weights are configured as $w_g = 1.0$ and $w_v = 2.5[s/m]$, ensuring that a typical human reaching motion balances equivalently with a stable gaze trigger. The final decision-making is supervised by the activation threshold $\tau$.

We executed an offline parametric sensitivity sweep of the continuous activation threshold $\tau \in \{0.60, 0.70, 0.80, 0.90\}$. An aggressive configuration ($\tau = 0.60$) maximizes temporal tracking responsiveness but renders the single-frame gating highly vulnerable to transient variations, causing premature latching and yielding a high frame-level FPR of 72.5\%. Conversely, an overly conservative threshold ($\tau = 0.90$) thoroughly suppresses frame-level sensory noise to an exceptionally low FPR of 9.8\%, but it severely restricts the proactive motion window, frequently missing object presentation phases and leading to interactive deadlocks. Our selected operational configuration of $\tau = 0.80$ marks the ideal value, securing an optimal balance by maximizing intent tracking sensitivity for smooth haptic transitions while strictly dampening the offline continuous frame-level FPR to 13.6\%.

\section{Discussions and Limitations}
\label{sec:discussions}
Our empirical evaluation yields several key insights and highlights open boundaries for future research. First, we observe that sensor-specific noise simulation is critical for cross-modal policy transfer; while frameworks like DepthCrafter~\cite{Hu2025depthcrafter} produce smooth geometry, our morphological noise simulation replicates the stochastic edge scattering characteristic of real depth sensors (e.g., RealSense L515), preventing the downstream network from overfitting to idealized synthetic structures. Second, by combining 6-DoF grasp annotations with timestamped intention labels, our pipeline enables a paradigm shift from hand-centric reactive observation to proactive, intention-aware human-robot collaboration. 

% Despite these advantages, certain limitations remain. Our current generation relies on re-injecting segmented characters into static backgrounds, which inherently constrains background environmental diversity~\cite{song2025mitty}. Furthermore, while our system-level evaluations across diverse human subjects quantitatively validate the physical safety and operational robustness of the proposed framework, a formalized human-in-the-loop qualitative user study was not conducted due to real-world robotic interaction safety constraints and strict institutional verification boundaries. Consequently, our empirical evidence primarily characterizes objective perception tracking profiles rather than subjective user comfort or psychological fluency. Future work will naturally explore user-centric haptic alignment surveys and subjective interaction naturalness grading to further refine the collaborative gating logic.

Despite these advantages, certain limitations remain. Our generation relies on re-injecting segmented characters into static backgrounds, constraining environmental diversity~\cite{song2025mitty}. Furthermore, due to robotic safety constraints and strict institutional verification boundaries, a formalized qualitative user study was not conducted. Consequently, our empirical evidence characterizes objective perception tracking profiles rather than subjective user comfort or psychological fluency. Future work will explore user-centric haptic alignment surveys and subjective naturalness evaluations to further refine the collaborative gating logic.

\end{document}